\documentclass{llncs}

\usepackage{llncsdoc}
\usepackage{times}
\usepackage{epsfig}
\usepackage{graphicx}
\usepackage{amsmath}
\usepackage{amssymb}
\usepackage{subfigure} 
\usepackage{color}
\usepackage{wrapfig}

\usepackage[pagebackref=true,breaklinks=true,letterpaper=true,colorlinks,bookmarks=false]{hyperref}

\begin{document}

\title{Detecting Synapse Location and Connectivity by Signed Proximity Estimation and Pruning with Deep Nets}

\author{Toufiq Parag\inst{1} \and Daniel Berger\inst{2} \and Lee Kamentsky\inst{1} \and Benedikt Staffler\inst{3} \and Donglai Wei\inst{1}  \and Moritz Helmstaedter\inst{3} \and 
Jeff W. Lichtman\inst{2} \and Hanspeter Pfister\inst{1}}

\institute{SEAS, Harvard University, Cambridge, MA 
 \and MCB, Harvard University, Cambridge, MA 
 \and Max Planck Institute for Brain Research, Frankfurt, Germany \\ 
 \email{email: toufiq.parag@gmail.com}}


\maketitle
\thispagestyle{empty}

\setcounter{footnote}{0}


\begin{abstract}

Synaptic connectivity detection is a critical task for neural reconstruction from Electron Microscopy (EM) data. Most of the existing algorithms for synapse detection do not identify the cleft location and direction of connectivity simultaneously. The few methods that computes direction along with contact location have only been demonstrated to work on either dyadic (most common in vertebrate brain) or polyadic (found in fruit fly brain) synapses, but not on both types. In this paper, we present an algorithm to automatically predict the location as well as the direction of both dyadic and polyadic synapses. The proposed algorithm first generates candidate synaptic connections from voxelwise predictions of signed proximity generated by a 3D U-net. A second 3D CNN then prunes the set of candidates to produce the final detection of cleft and connectivity orientation. Experimental results demonstrate that the proposed method outperforms the existing methods for determining synapses in both rodent and fruit fly brain\footnote{\scriptsize Code at: \url{https://github.com/paragt/EMSynConn}}.

\end{abstract}

\section{Introduction}

Connectomics has become a fervent field of study in neuroscience and computer vision recently. The goal of EM connectomics is to reconstruct the neural wiring diagram from Electron Microscopic (EM) images of animal brain. Neural reconstruction of EM images consists of two equally important tasks: (1) trace the anatomical structure of each neuron by labeling each voxel within a cell with a unique id; and (2) find the location and direction of synaptic connections among multiple cells.

The enormous amount of EM volume emerging from a tiny amount of tissue constrains any subsequent analysis to be performed (semi-) automatically to acquire a comprehensive knowledge within a practical time period~\cite{jain10opinion}\cite{helmstaedter15}. Discovering the anatomical structure entails a 3D segmentation of EM volume. Numerous studies have addressed this task with many different approaches, we refer interested readers to~\cite{funke17arxiv}\cite{lee17superhuman}\cite{januszewski16flood}\cite{parag2015ICCV} \cite{liu16eccv}\cite{parag14}\cite{beier2017MulticutBA} for further details. In order to unveil the connectivity, it is necessary to identify the locations \emph{and} the direction of synaptic communications between two or more cells. Resolving the location of synaptic contact is crucial for neurobiological reasons, and, because the strength of connection between two cells is determined by the number of times they make a synaptic contact. The direction of the synaptic contact reveals the direction of information flow from presynaptic to postsynaptic cells. By defining the edges, the location and connectivity orientation of synapses complete the directed network of neural circuitry that a neural reconstruction seek to discover. In fact, discovering synaptic connectivity was one of the primary reasons to employ the immensely complex and expensive apparatus of electron microscopy for connectomics in the first place. Other imaging modalities (e.g., light microscopy) are either limited by their resolution or by a conclusive and exhaustive strategy (e.g., using reagents) to locate synapses~\cite{morgan13whynot}\cite{denk12}\cite{lichtman14bigdata}. 

In terms of complexity, identification of neural connectivity is as challenging as tracing the neurons~\cite{dorkenwald17syconn}. With rapid and outstanding improvement in automated EM segmentation in recent years, detection of synaptic connectivity may soon become a bottleneck in the overall neural reconstruction process~\cite{staffler17synem}. Although fewer in number when compared against those in neurite segmentation, there are past studies on synaptic connectivity detection; we mention some notable works in the relevant literature section below. Despite many discernible merits of previous works, very few of them aim to identify both the location and direction of synaptic junctions. Among these few methods, namely by~\cite{kreshuk15miccai}\cite{dorkenwald17syconn}\cite{staffler17synem}, none of them have been shown to be generally applicable on different types of synapses typically found on different species of animals, e.g., dyadic connections in vertebrate (mouse, rat, zebrafinch, etc.) and polyadic connections in non-vertebrate (fruit fly) brain\footnote{\scriptsize Although, there are examples of polyadic connections in mouse cerebellum between mossy fibers and granule cells}. Apart from a few, the past approaches do not benefit from the advantages deep (fully) convolutional networks offer. Use of hand crafted features could stifle the utility of a method on widely divergent EM volumes collected from different animals with different tissue processing and imaging techniques. 

In this paper, we propose a general method to automatically detect both the 3D location and direction of both dyadic (vertebrate) and polyadic (fruit fly) synaptic contacts. The proposed algorithm is designed to predict the proximity (and polarity, as we will explain later) of every 3D voxel to a synaptic cleft using a deep fully convolutional network, namely a U-net~\cite{ronneberger15unet}. A set of putative locations, along with their connection direction estimates, are computed given a segmentation of the volume and the voxelwise prediction from the U-net. A second stage of pruning, performed by a deep convolutional network, then trims the set of candidates to produce the final prediction of 3D synaptic cleft locations and the corresponding directions. 

The use of CNNs makes the proposed approach generally applicable to new data without the need for feature selection. Estimation of the location and connectivity at both voxel and segment level improves the accuracy but do not require any additional annotation effort (no need for labels for other classes such as vesicles). We show that our proposed algorithm outperforms the existing approaches for synaptic connectivity detection on both vertebrate and fruit fly datasets. Our evaluation measure (Section~\ref{S:EXPT}), which is critical to assess the actual performance of a synapse detection method, has also been confirmed by a prominent neurobiologist to correctly quantify actual mistakes on real dataset.

\subsection{Relevant Literature:} Initial studies on automatic synaptic detection focused on identifying the cleft location by classical machine learning/vision approaches using pre-defined features~\cite{kreshuk11plos}\cite{becker13syn}\cite{kreshuk14plos} \cite{plaza14syn}\cite{jagadeesh14}\cite{huang14syn}. These algorithms assumed subsequent human intervention to determine the synaptic partners given the cleft predictions. Roncal et al.~\cite{roncal14vesicle} combine the information provided by membrane and vesicle predictions with a CNN (not fully convolutional) and apply post-processing to locate synaptic clefts.  To establish the pre- and post-synaptic partnership, ~\cite{huang16} augmented the synaptic cleft detection with a multi-layer perceptron operating on hand designed features. On the other hand, Kreshuk et al.~\cite{kreshuk15miccai} seek to predict vesicles and clefts for each neuron boundary by a random forest classifier (RF) and then aggregate these predictions with a CRF to determine the connectivity for polyadic synapses in fruit fly. Dorkenwald et al.~\cite{dorkenwald17syconn} utilize multiple CNNs, RFs to locate synaptic junctions as well as vesicles, mitochondria, and to decide the dyadic orientation in vertebrate EM data. SynEM~\cite{staffler17synem} attempts to predict connectivity by classifying each neuron boundary (interfaces) to be synaptic or not using a RF classifier operating on a confined neighborhood and has been shown to perform better than~\cite{dorkenwald17syconn} in terms of connectivity detection.

\section{Method}
The proposed method is designed to first predict both the location and direction of synaptic communication at the voxel level. Section~\ref{S:PIXEL} illustrates how this is performed by training a deep encoder-decoder type network, namely the U-net, to compute the proximity and direction of connection with respect to synapses at every voxel. The voxelwise predictions are clustered together after discretization and matched with a segmentation to establish synaptic relations between pairs of segment ids. Afterwards, a separate CNN is trained to discard the candidates that do not correspond to an actual synaptic contact, both in terms of location and direction, as described in Section ~\ref{S:CANDIDATE_PRUNE}.

\subsection{Voxelwise Synaptic Cleft and Partner Detection}\label{S:PIXEL}
\begin{figure*}[t]
\vspace{-0.2cm}
\begin{center}
\vspace{-0.2cm}
\subfigure[\scriptsize Traditional annotation 2D]{\includegraphics[width=0.3\columnwidth, height=0.25\columnwidth]{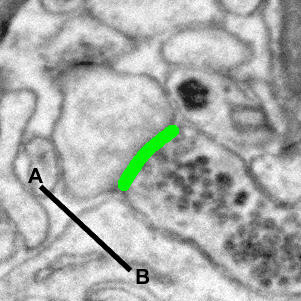}\label{F:PREV_LABEL}\vspace{-0.6cm}}
\subfigure[\scriptsize Proposed label 2D]{\includegraphics[width=0.3\columnwidth, height=0.25\columnwidth]{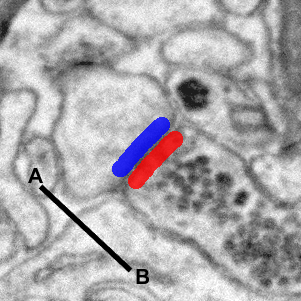}\label{F:PROPOSED_LABEL}\vspace{-0.6cm}}
\subfigure[\scriptsize Proposed target function 2D]{\includegraphics[width=0.3\columnwidth, height=0.25\columnwidth]{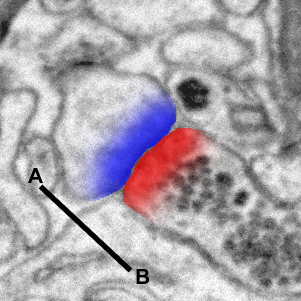}\label{F:FUNC_APPROX}\vspace{-0.6cm}}
\subfigure[\scriptsize Traditional annotation 1D]{\vspace{-0.6cm}\includegraphics[width=0.3\columnwidth, height=0.22\columnwidth]{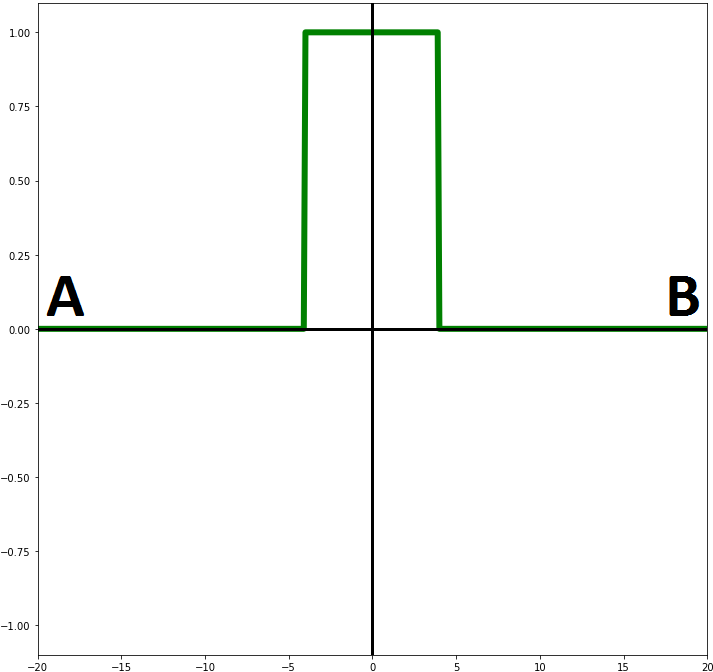}\label{F:PREV_LABEL1D}}
\subfigure[\scriptsize Proposed annotation 1D]{\includegraphics[width=0.3\columnwidth, height=0.22\columnwidth]{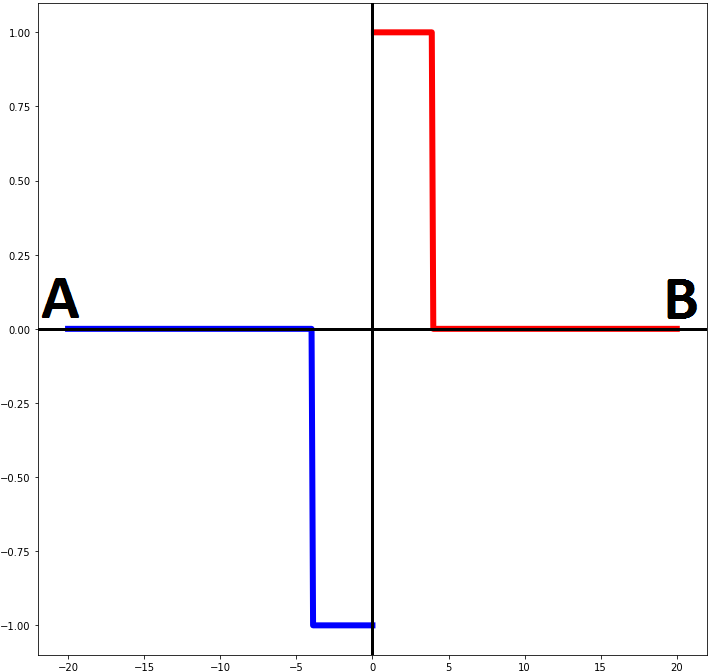}\label{F:PROPOSED_LABEL1D}}
\subfigure[\scriptsize Proposed target function 1D]{\includegraphics[width=0.3\columnwidth, height=0.22\columnwidth]{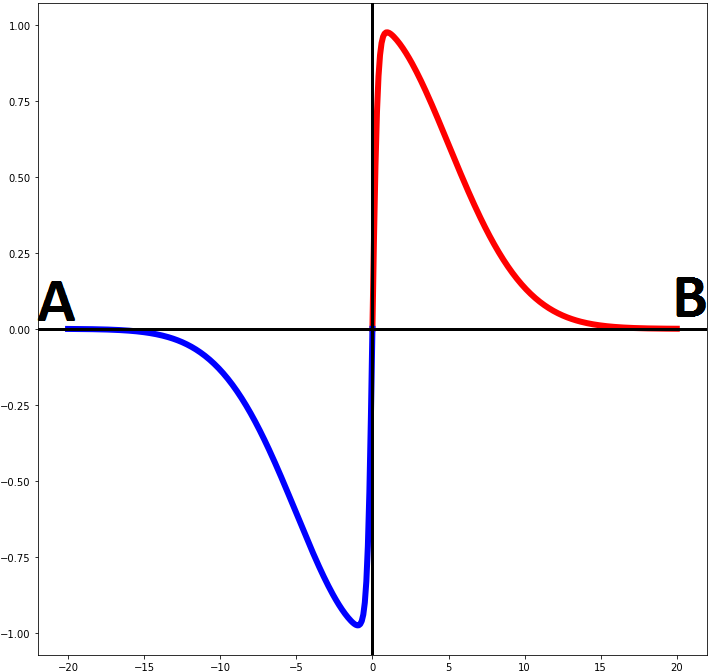}\label{F:FUNC_APPROX1D}}
\vspace{-0.4cm}
\caption{\scriptsize The traditional and proposed annotation of synapses and the signed proximity function that the proposed method estimates are shown in 2D in \ref{F:PREV_LABEL}, \ref{F:PROPOSED_LABEL}, \ref{F:FUNC_APPROX} and in 1D in \ref{F:PREV_LABEL1D}, \ref{F:PROPOSED_LABEL1D}, \ref{F:FUNC_APPROX1D} respectively for illustration purposes. The U-net learns the signed proximity functions in 3D.}\label{F:ANNOTATION}
\end{center}
\vspace{-0.4cm}
\end{figure*}

In order to learn both the position and connection orientation of a synapse, the training labels for voxels are modified slightly from the traditional annotation. It is the standard practice to demarcate only the synaptic cleft with a single strip of id, or color, as the overlaid color shows in Figure~\ref{F:ANNOTATION}\subref{F:PREV_LABEL}. In contrast, the proposed method requires the neighborhood of the pre- and post-synaptic neurons at the junction of synaptic expression to be marked by two ids or colors as depicted in Figure~\ref{F:ANNOTATION}\subref{F:PROPOSED_LABEL}. To distinguish the partners unambiguously, these ids can follow a particular pattern, e.g., pre-synaptic partners are always marked with odd id and post-synaptic partners are annotated with even ids. Such annotations inform us about both the location and direction of a connection with practically no increase in annotation effort. Note that, although we explain and visualize the labels in 2D and in 1D for better understanding, our proposed method learns a function of the 3D annotations.   

For voxelwise prediction of position and direction, ids of all pre- and post-synaptic labels (red and blue in Figure~\ref{F:ANNOTATION}\subref{F:PROPOSED_LABEL}) are converted to 1 and -1 respectively; all remaining voxels are labeled with 0. However, our approach \emph{does not} directly learn from the discrete labels presented in~\ref{F:ANNOTATION}\subref{F:PROPOSED_LABEL}. Instead, we attempt to learn a smoother version of the discrete labels, where the transition from 1 and -1 to 0 occurs gradually, as shown in Figure~\ref{F:ANNOTATION}\subref{F:FUNC_APPROX}. The dissimilarities among these three types of annotations can be better understood in 1D. The Figures~\ref{F:ANNOTATION}\subref{F:PREV_LABEL1D}, \subref{F:PROPOSED_LABEL1D}, and~\subref{F:FUNC_APPROX1D} plot the labels perpendicular to the black line drawn underneath the labels in images in the top row of Figure~\ref{F:ANNOTATION}. The proposed approach attempts to learn a 3D version of the smooth function in Figure~\ref{F:ANNOTATION}\subref{F:FUNC_APPROX} and~\subref{F:FUNC_APPROX1D}. Effectively, this target function computes a signed proximity from the synaptic contact, where the sign indicates the connectivity orientation and the absolute function value determines the closeness or affinity of any voxel to the nearest cleft. Mathematically, this function can be formulated as
\vspace{-0.2cm}
\begin{equation}
\small \text{proximity}_\text{signed}(x) = \biggl \{\exp {- d(x)^2 \over {2 \sigma^2}} \biggr \} ~ \biggl \{ {2 \over {1 + \exp( - \alpha~ d(x))}} - 1 \biggr \}, ~ 
\vspace{-0.2cm}
\end{equation}
where $d(x)$ is the signed distance between voxel at $x$ and the synaptic cleft, and $\alpha$ and $\sigma$ are parameters that control the smoothness of transition. We solve a regression problem using a 3D implementation of U-net with linear final layer activation function to approximate this function.  There are multiple motivations to approximate a smooth signed proximity function. A smooth proximity function as a target also eliminates the necessity of estimating the abrupt change near the annotation boundary and therefore assists the gradient descent to approach a more meaningful local minimum.  Furthermore, some recent studies have suggested that a smooth activation function is more useful than its discrete counterparts for learning regression~\cite{hou17smooth}\cite{mobahi16}.  

Our 3D U-net for learning signed proximity has an architecture similar to the original U-net model in~\cite{ronneberger15unet}. The network has a depth of 3 where it applies two consecutive $3 \times 3 \times 3$ convolutions at each depth and utilizes parametric leaky ReLU~\cite{he15prelu} activation function. The activation function in the final layer is linear. The input and output of the 3D U-net are $316\times 316\times 32$ grayscale EM volumes and $228\times 228\times 4$ proximity values, respectively. A weighted mean squared error loss is utilized to learn the proximity values during training.

\subsection{Candidate Generation and Pruning}\label{S:CANDIDATE_PRUNE}
\begin{figure*}[t]
\vspace{-0.2cm}
\begin{center}
\vspace{-0.2cm}
\subfigure[\scriptsize EM image]{\includegraphics[width=0.24\columnwidth, height=0.23\columnwidth]{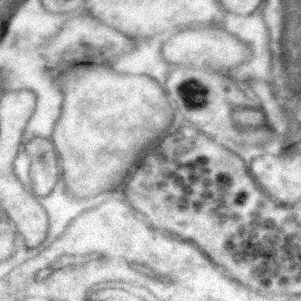}\label{F:IMAGE}\vspace{-0.4cm}}
\subfigure[\scriptsize Proximity prediction]{\includegraphics[width=0.24\columnwidth, height=0.23\columnwidth]{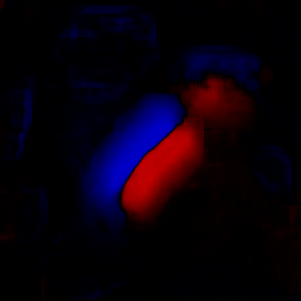}\label{F:PRED}\vspace{-0.4cm}}
\subfigure[\scriptsize Segmentation mask]{\includegraphics[width=0.24\columnwidth, height=0.23\columnwidth]{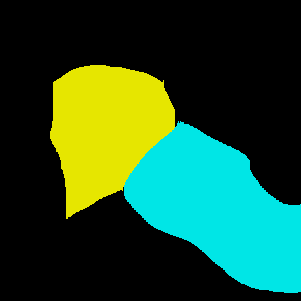}\label{F:SEG}\vspace{-0.4cm}}
\subfigure{\raisebox{8mm} {\includegraphics[width=0.25\columnwidth, height=0.07\columnwidth]{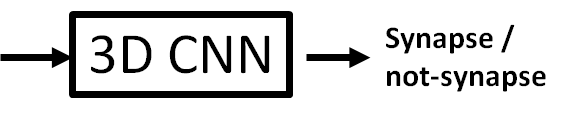}}}
\vspace{-0.4cm}
\caption{\scriptsize Candidate pruning by 3D CNN. The EM image,  proximity prediction and segmentation mask on one section of input subvolume are shown in \subref{F:IMAGE}, \subref{F:PRED},  \subref{F:SEG} respectively. The cyan and yellow segmentation masks are provided as separate binary masks, shown here in one image \subref{F:SEG} to save space.}\label{F:PRUNE}
\end{center}
\vspace{-0.4cm}
\end{figure*}

For computing putative pairs of pre- and post-synaptic partners, we first threshold the signed proximity values at an absolute value of $\tau$ and compute connected components for pre and post-synaptic sites separately. Given a segmentation $S$ for the EM volume, every pre-synaptic connected component $e$ is paired with one or more segment $s_{i_e} \in S,~ i_e = 1, \dots, m_e $ based on a minimum overlap size $\omega$ to form pre-synaptic site candidates $T_{e,{i_e}}$. Similarly, post-synaptic site candidates $T_{o,{j_o}}$ are generated by associating each post-synaptic connected component $o$ is with a set of segments $ s_{j_o} \in S,~ j_o = 1, \dots, n_o$. The set $\cal C$ of candidate pairs of synaptic partners are computed by pairing up pre-synaptic candidate $T_{e,{i_e}}$  with post-synaptic $T_{o,{j_o}}$ wherever segment $s_{j_o}$ is a neighbor of $s_{i_e}$, i.e., $s_{i_e}$ shares a boundary with segment $s_{j_o}$.
\vspace{-0.2cm}
\begin{equation}
\small {\cal C} = \Bigl \{ \{ T_{e,{i_e}}, T_{o,{j_o}} \}~|~ s_{j_o} \in \text{Nbr}(s_{i_o}), ~\forall {e, o, i_e, j_o} \Bigr \}. 
\vspace{-0.2cm}
\end{equation}
A 3D CNN is utilized to distinguish  the correct synaptic partner pairs from the false positive candidates, i.e., to produce a binary decision for each candidate. The groundtruth labels for training this second convolutional network were computed by matching the segmentation $S$ with the groundtruth segmentation of the volume $G$. The pruning network is constructed with 5 layer convolutions of size $3 \times 3 \times 3$ followed by two densely connected layers. The input to the 3D deep convolutional pruning (or trimming) network comprises $160\times 160\times 16$ subvolumes of  the grayscale EM image, the predicted signed proximity values from 3D U-net and the segmentation masks of $s_{i_e}, s_{j_o}$, extracted from a 3D window centered at the closest point between the connected components $e$ and $o$, as shown in Figure~\ref{F:PRUNE}. 

It is worth mentioning here that we have contemplated the possibility of combining the voxelwise network and the candidate trimming network to facilitate end-to-end training, but did not pursue that direction due to the difficulty in formulating a differentiable operation to transform voxelwise signed proximities to region wise candidates. We have, however, attempted to employ a region proposal network based method, in particular the mask R-CNN~\cite{he2017maskrcnn}, to our problem. The proposal generating network of mask R-CNN method resulted in a low recall rate for locating the  synapses in our experiments (leading to low recall in the final result after trimming). We observed Mask R-CNNs to struggle with targets with widely varying size in our dataset. Furthermore, we had difficulty in merging many proposals~\cite{he2017maskrcnn} produced for one connection, leading to lower final precision rate as well. 


\section{Experiments and Results}\label{S:EXPT}
The deep nets for this work were implemented in Theano with Keras interface\footnote{\scriptsize Code at: \url{https://github.com/paragt/EMSynConn}}. For both the network training, we used rotation and flip in all dimensions for data augmentation. For the candidate trimming network, we also displaced the center of the window by a small amount to augment the training set. The parameters for the target signed proximity and the candidate overlap calculation remained the same as $\alpha = 5, \tau = 0.3~\text{(absolute value)}, \omega = 100$. The parameter  $\sigma$ was set to $10$ for the mouse and rat dataset but $14$ for the TEM fly data to account for the difference in z-resolution. 

\noindent \textbf{Evaluation:} It is critical to apply the most biologically meaningful evaluation formula in order to correctly assess the performance of any given method. Distance based methods~\cite{staffler17synem}\cite{becker13syn}, for example, are unrealistically tolerant to false positive detections nearby. On the other hand, pixelwise error computation~\cite{cremi} is more stringent than necessary for extracting the wiring diagram -- two detections with $50\%$ and $60\%$ pixelwise overlap need not be penalized differently for connectomics purposes. Measures computed solely on overlap~\cite{roncal14vesicle} becomes ambiguous when one prediction overlaps two junctions. We resolve this ambiguity by considering a detection be correct only if it overlaps with the span of synaptic expression (as delineated by an expert) \emph{and} connects two cells with correct synaptic orientation, i.e., pre and post-synaptic partners. All the precision and recall values reported in the experiments on rat (Section~\ref{S:RAT}) and mouse (Section~\ref{S:MOUSE}) data are computed with this measure.

 
\subsection{Rat Cortex}\label{S:RAT}
Our first experiment was designed to determine the utility of the two stages, i.e., voxelwise prediction and candidate set pruning, of the  proposed algorithm. The EM images we used in this experiment were collected from rat cortex using SEM imaging at resolution of $4\times 4\times 30$nm. We used a volume of $97$ images to train the 3D U-net and validated on a different set of $120$ images. The candidate pruning CNN was trained on $97$ images and then fine tuned on the $120$ image dataset. For testing we used a third volume of $145$ sections. The segmentation used to compute the synaptic direction was generated either by the method of~\cite{parag17arxiv}.

\begin{figure*}[t]
\vspace{-0.2cm}
\begin{center}
\vspace{-0.2cm}
\subfigure[\scriptsize Rat cortex PR curve]{\includegraphics[width=0.43\columnwidth, height=0.32\columnwidth]{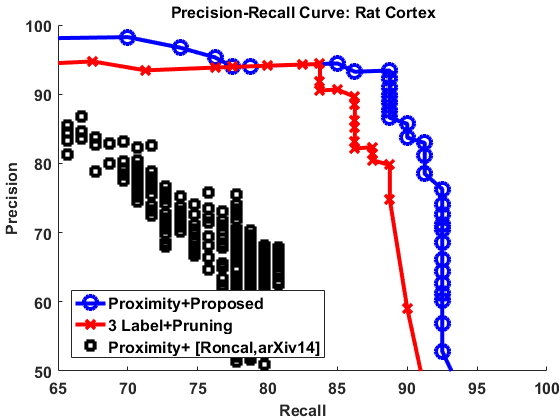}\label{F:ECS_COMP}\vspace{-0.4cm}}
\subfigure[\scriptsize Mouse cortex(SNEMI) PR curve]{\includegraphics[width=0.43\columnwidth, height=0.32\columnwidth]{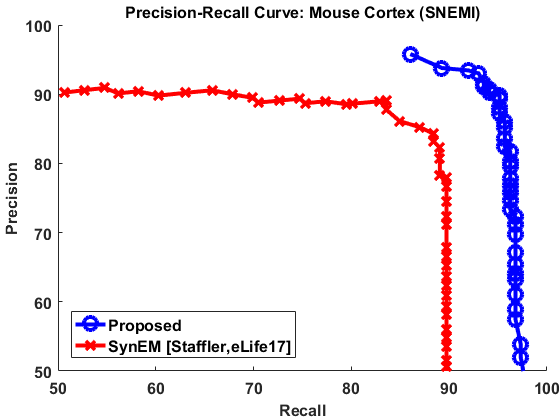}\label{F:KASTHURI_COMP}\vspace{-0.4cm}}
\vspace{-0.5cm}
\caption{\scriptsize Precision recall curves for \emph{synapse location and connectivity} for rat cortex~\subref{F:ECS_COMP} and mouse cortex (SNEMI) \subref{F:KASTHURI_COMP} experiments. Plot \subref{F:ECS_COMP} suggests combining voxelwise signed proximity prediction with pruning performs significantly better than the versions that replaces one of these components with alternative strategies. Plot \subref{F:KASTHURI_COMP} indicates significant improvement achieved over performance of~\cite{staffler17synem} on the same dataset.}\label{F:VERTEBRATE_COMP}
\end{center}
\vspace{-0.6cm}
\end{figure*}

Figure~\ref{F:VERTEBRATE_COMP}\subref{F:ECS_COMP} compares the precision recall curves for detecting both location and connectivity with two variants.  1) 3 Label + pruning - where the proximity approximation is replaced by 3-class classification among pre-, post-synaptic, and rest of the voxels (Figure~\ref{F:ANNOTATION}\subref{F:PROPOSED_LABEL}). 2) Proximity + [Roncal, arXiv14] - where the proposed pruning network is replaced by VESICLE~\cite{roncal14vesicle} style post-processing. For the proposed (blue o) and 3 Label + pruning (red x) technique, each point on the plot correspond to a threshold on the prediction of the 3D trimming CNN. For the Proximity + [Roncal, arXiv14] technique (black o), we varied several parameters of the VESICLE post-processing.

This experiment suggests that the pruning network is substantially more effective than morphological post-processing~\cite{roncal14vesicle}. The proposed signed proximity approximation yields $3\%$ more true positives in the initial candidate set than those generated by the multiclass prediction. As discussed earlier in Section~\ref{S:PIXEL}, a smooth target function places the focus on learning difficult examples by removing the necessity to learn the sharp boundaries. Empirically, we have noticed the training procedure to spend a significant number of iterations to infer the sharp boundary of a discrete label like Figure\ref{F:ANNOTATION}\subref{F:PROPOSED_LABEL} and \subref{F:PROPOSED_LABEL1D}. Furthermore, we observed that the wider basin of prediction can identify more true positives and offers more information for the following pruning stage to improve the F-score of our method to $91.03\%$ as opposed $87.5\%$ of the 3 Label + pruning method. 

\subsection{Mouse Cortex data (SNEMI)}\label{S:MOUSE}
We experimented next on the SEM dataset from Kasthuri et.al.~\cite{kasthuri15cell} that was used for the SNEMI challenge~\cite{snemi} to compare our performance with that of SynEM~\cite{staffler17synem} (which was shown to outperform~\cite{dorkenwald17syconn}). The synaptic partnership information was collected from the authors of~\cite{kasthuri15cell} to compute the signed proximities for training the 3D U-net. Similar to the rat cortex experiment, we used 100 sections for training the voxelwise U-net and candidate pruning CNN and used 150 sections for validation of the voxelwise proximity U-net. The test dataset consists of 150 sections of size $1024\times 1024$ that is referred to as AC3 in~\cite{staffler17synem}. The segmentation used to compute the synaptic direction was generated either by the method of~\cite{parag17arxiv}. 

\begin{figure*}[h]
\vspace{-0.2cm}
\begin{center}
\vspace{-0.2cm}
\subfigure[\scriptsize GT annotation]{\includegraphics[width=0.3\columnwidth, height=0.24\columnwidth]{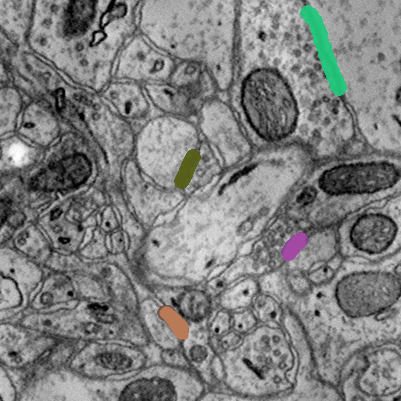}\label{F:GT}\vspace{-0.4cm}}
\subfigure[\scriptsize SynEM pred]{\includegraphics[width=0.3\columnwidth, height=0.24\columnwidth]{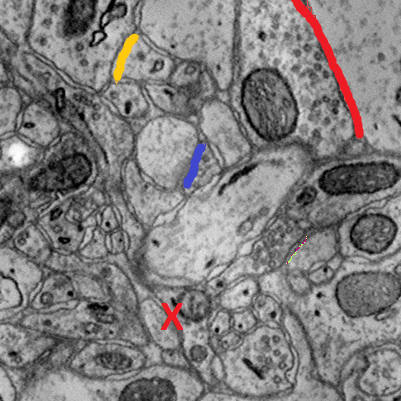}\label{F:SYNEM}\vspace{-0.4cm}}
\subfigure[\scriptsize Proposed prediction]{\includegraphics[width=0.3\columnwidth, height=0.24\columnwidth]{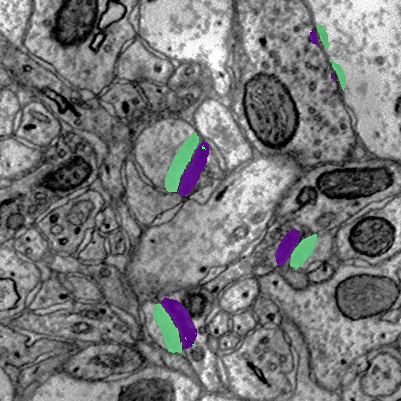}\label{F:PROPOSED_KASTHURI}\vspace{-0.4cm}}
\vspace{-0.5cm}
\caption{\scriptsize Qualitative results on mouse cortex data~\cite{kasthuri15cell}. Left to right, groundtruth annotation, output of SynEM~\cite{staffler17synem} (red x marks missed location), and that of the proposed method. In \subref{F:PROPOSED_KASTHURI}, purple and green indicates pre and post-synaptic partner respectively. }\label{F:KASTHURI_QUAL}
\end{center}
\vspace{-0.7cm}
\end{figure*}

The authors of~\cite{staffler17synem} have kindly provided us with their results on this dataset. We read off the cleft detection and the connectivity estimation from their result and computed the error measures as explained in Section~\ref{S:EXPT}. The precision recall curve for detecting both cleft and connectivity is plotted in Figure~\ref{F:VERTEBRATE_COMP}\subref{F:KASTHURI_COMP}. In general, the SynEM method performs well overall, but produces significantly higher false negative rates than the proposed method. Among the synapses it detects, SynEM assigns the pre and post-synaptic partnership very accurately, although it used the actual segmentation groundtruth for such assignment whereas we used a segmentation output. Visual inspection of the detection performances also verifies the lower recall rate of~\cite{staffler17synem} than ours. In Figure~\ref{F:KASTHURI_QUAL} we show the groundtruth, the cleft estimation of SynEM (missed connection marked with red x) and the connected components corresponding to the predictions of the proposed method, both computed at the largest F-score. The direction of synaptic connection in image of Figure~\ref{F:KASTHURI_QUAL}\subref{F:PROPOSED_KASTHURI} is color coded -- purple and green indicates pre and post-synaptic partners respectively.   

\subsection{Fruit Fly data (CREMI)}\label{S:CREMI}
Our method was applied on the 3 TEM volumes (resolution $4\times 4\times 40$nm) of the CREMI challenge~\cite{cremi}. We annotated the training labels for synaptic partners for all 3 volumes given the synaptic partner list provided on the website. All the experimental settings for this experiment remain the same as other except those mentioned in Section~\ref{S:EXPT}. Out of the 125 training images, we used the first 80 for training and the remaining images for validation. The segmentation used to compute the synaptic direction was generated either by the method of~\cite{funke17arxiv}. 

The performance is only measured in terms of synaptic partner identification task,  as the pixelwise cleft error measure is not appropriate for our result (refer to the output provided in Figure~\ref{F:KASTHURI_QUAL}\subref{F:PROPOSED_KASTHURI}). At the time of this submission the our method, which is identified as HCBS on CREMI leaderboard, holds the 2nd place overall (error differenece with the first is 0.002) and performed better than both variants of~\cite{kreshuk15miccai}.
\begin{table}[h]
\scriptsize 
\vspace{-0.4cm}
\caption{\scriptsize Result on CREMI data, lower is better}
\label{T:CREMI}
\centering
\vspace{-0.2cm}
\begin{tabular}{c c | c c c} 
 \hline
 Method & Submission & CREMI score & FP & FN \\  
 \hline
 HCBS(proposed) & Tr66\_80K & $0.449$ & $223.000$ & $286.667$  \\
 IAL~\cite{kreshuk15miccai} & PSP\textunderscore unar & $0.461$ & $266.667$ & $281.000$ \\ 
 IAL~\cite{kreshuk15miccai} & PSP\textunderscore full & $0.464$ & $187.333$ & $310.000$ \\
 \hline
\end{tabular}
\vspace{-0.4cm}
\end{table}

\section{Conclusion} 
We propose a general purpose synaptic connectivity detector that locates the location and direction of a synapse at the same time. Our method was designed to work on both dyadic and polyadic synapses without any modification to its component techniques. The utilization of deep CNNs for learning location and direction of synaptic communication enables it to be directly applicable to any new dataset without the need for manual selection of features. Experiments on multiple datasets suggests the superiority of our method on existing algorithms for synaptic connectivity detection. One straightforward extension of the proposed two stage method is to enhance the candidate pruning CNN to distinguish between excitatory and inhibitory synaptic connections by adopting a 3-class classification scheme.
{\scriptsize
\bibliographystyle{splncs}
\bibliography{arxiv18synapse}
}

\end{document}